\documentclass[conference]{IEEEtran}

\IEEEoverridecommandlockouts                              %

\usepackage{xcolor}
\usepackage{amsmath}
\usepackage{amssymb}
\usepackage{mathtools}

\usepackage{amsthm}
\usepackage{hyperref}
\usepackage{xspace}
\usepackage[ruled,vlined,linesnumbered]{algorithm2e}
\usepackage{soul}
\usepackage{graphicx}
\usepackage[caption=false,font=footnotesize]{subfig}

\newtheorem{prop}{Proposition}

\newtheorem*{remark}{Remark}

\usepackage{enumitem}
\newcommand{\risk}{\mathcal R}
\newcommand{\pmp}{Pontryagin's minimum principle\xspace}
\newcommand{\thmFw}{Proposition~\ref{thm:eqFrank}\xspace}
\newcommand{\gJ}{\ensuremath{\nabla_{w_n} \hat{J}(w_0, w_1, \dots, w_{N-1})}\xspace}
\newcommand{\gf}{\ensuremath{\nabla_{w_n} f(x_n, w_n)}\xspace}
\newcommand{\gHnw}{\ensuremath{\nabla_{w_n} H(x_n,p_{n+1},w_n)}\xspace}

\newcommand{\pphi}{\ensuremath{p_{n+1}^\top  \hat{\phi}(x_n)^\top }\xspace }
\newcommand{\pphisig}{\ensuremath{p_{n+1}^\top  \hat{\phi}(x_n)^\top \Sigma^{1/2}}\xspace }
\newcommand{\pphisigt}{\ensuremath{(\Sigma^{1/2})^\top   \hat{\phi}(x_n) p_{n+1}}\xspace }

\newcommand{\lip}{linear-in-parameter\xspace}
\newcommand{\rbfk}{Gaussian RBF kernel\xspace}

\newcommand{\Jws}{\ensuremath{ J(w_0, w_1, \dots, w_{N-1})}\xspace}
\newcommand{\fxuw}{\ensuremath{f(\mathbf{x}_n, \mathbf{w}_n)}\xspace}
\newcommand{\algpmp}{Algorithm~\ref{alg:pmp}\xspace}
\newcommand{\ws}{\ensuremath{(w_0^*,w_1^*,\dots ,w_{N-1}^*)}\xspace}
\newcommand{\ps}{\ensuremath{(p_0^*,p_1^*,\dots ,p_{N}^*)}\xspace}
\newcommand{\xs}{\ensuremath{(x_0^*,x_1^*,\dots ,x_{N}^*)}\xspace}

\newcommand{\wc}{worst-case\xspace}
\newcommand{\fw}{Frank-Wolfe\space}
\newcommand{\cgd}{conditional gradient method}

\newcommand{\wset}{\ensuremath{\mathcal W}\xspace} %
\DeclareMathOperator*{\argmin}{arg\,min}
\DeclareMathOperator*{\argmax}{arg\,max}

\newcommand{\hip}[2]{\langle {#1, #2} \rangle _{\mathcal H}}
\renewcommand{\mathbf}[1]{#1}
\renewcommand{\boldsymbol}{}

\begin{document}
\title{\LARGE \bf 
Shallow Representation is Deep: Learning {{Uncertainty-aware and Worst-case Random Feature Dynamics}}
}

\author{Diego Agudelo-España, Yassine Nemmour, Bernhard Schölkopf, Jia-Jie Zhu%
\thanks{The authors are with the Empirical Inference Department, Max Planck Institute for Intelligent Systems, T\"ubingen, Germany
        {\tt\small \{first.last\}@tuebingen.mpg.de}}%
}

\maketitle
\thispagestyle{empty}
\pagestyle{empty}

\begin{abstract}
Random features is a powerful universal function approximator that inherits the theoretical rigor of kernel methods and can scale up to modern learning tasks.
This paper views uncertain system models as unknown or uncertain smooth functions in universal reproducing kernel Hilbert spaces.
By directly approximating the one-step dynamics function using random features with uncertain parameters, which are equivalent to a shallow Bayesian neural network,
we then view the whole dynamical system as a multi-layer neural network.
Exploiting the structure of Hamiltonian dynamics, we show that finding worst-case dynamics realizations using \pmp is equivalent to performing the \fw algorithm on the deep net.
Various numerical experiments on dynamics learning showcase the capacity of our modeling methodology.
\end{abstract}
\IEEEpeerreviewmaketitle

\section{Introduction}
\label{sec:intro}
An appropriate dynamics model is at the core of many learning-based control methods.
The problem of learning dynamical systems can be formulated as a function approximation task in numerical analysis.
Mathematically, we wish to solve the optimization problem
$
\min_{f\in \mathcal H} \| f - f^*\|,
$
where $f^*$ is the (unknown) true dynamics model of the underlying system and $\|\cdot\|$ some criteria such as function norms.
In this problem, the function class $\mathcal H$ plays a central role.
It can be chosen as, e.g., deep neural networks (DNN), Gaussian processes (GP).
As an example, let us consider a simplified optimal control problem (OCP)
\begin{equation}
	\label{eq:ocpIntro}  
	\min_{u_0,\dots,u_N} \sum_{t=1}^N c(x_t, u_t)\ \  \text{ subject to }{x}_{n+1} =  f({x}_n, {u}_n),x_0 = \hat{p} ,
\end{equation}
where $n=0, ..., N-1$.
Ideally, a learned dynamics model $f$ should allows us to reason about the worst-case objective value of $ \sum_{t=1}^N c(x_t, u_t)$ under model uncertainty (i.e., the highest plausible cost).

In the current literature, DNN are often celebrated as the most expressive \emph{universal function approximator}. Numerous works have considered DNN dynamics models in the learning-based control tasks, \textsf{}e.g., \cite{claveraLearningAdaptDynamic2019,picheNonlinearModelPredictive2000}.
However, despite efforts to propagate uncertainty under DNN dynamics (see \cite{chuaDeepReinforcementLearning2018} and references therein),
typical methods still rely on exhaustive simulations which struggle to be used with reliable numerical methods for robust and stochastic control.

On the other hand, GP dynamics models have enjoyed great success in learning-based control due to its uncertainty-aware nature.
For example,
many authors such as \cite{deisenrothPILCOModelbasedDataefficient2011,hewingCautiousModelPredictive2020} exploited the moment-matching of Gaussian distributions for uncertainty propagation.
Since moment-matching inevitably weakens any robustness guarantees,
the authors of \cite{kollerLearningbasedModelPredictive2018} proposed an over-approximation scheme.
However, it was pointed out that their method leads to \emph{over-conservatism}; see, e.g., \cite{lewSamplingbasedReachabilityAnalysis2020}.

\begin{figure}[t!]
    \centering
    \includegraphics[width=\columnwidth]{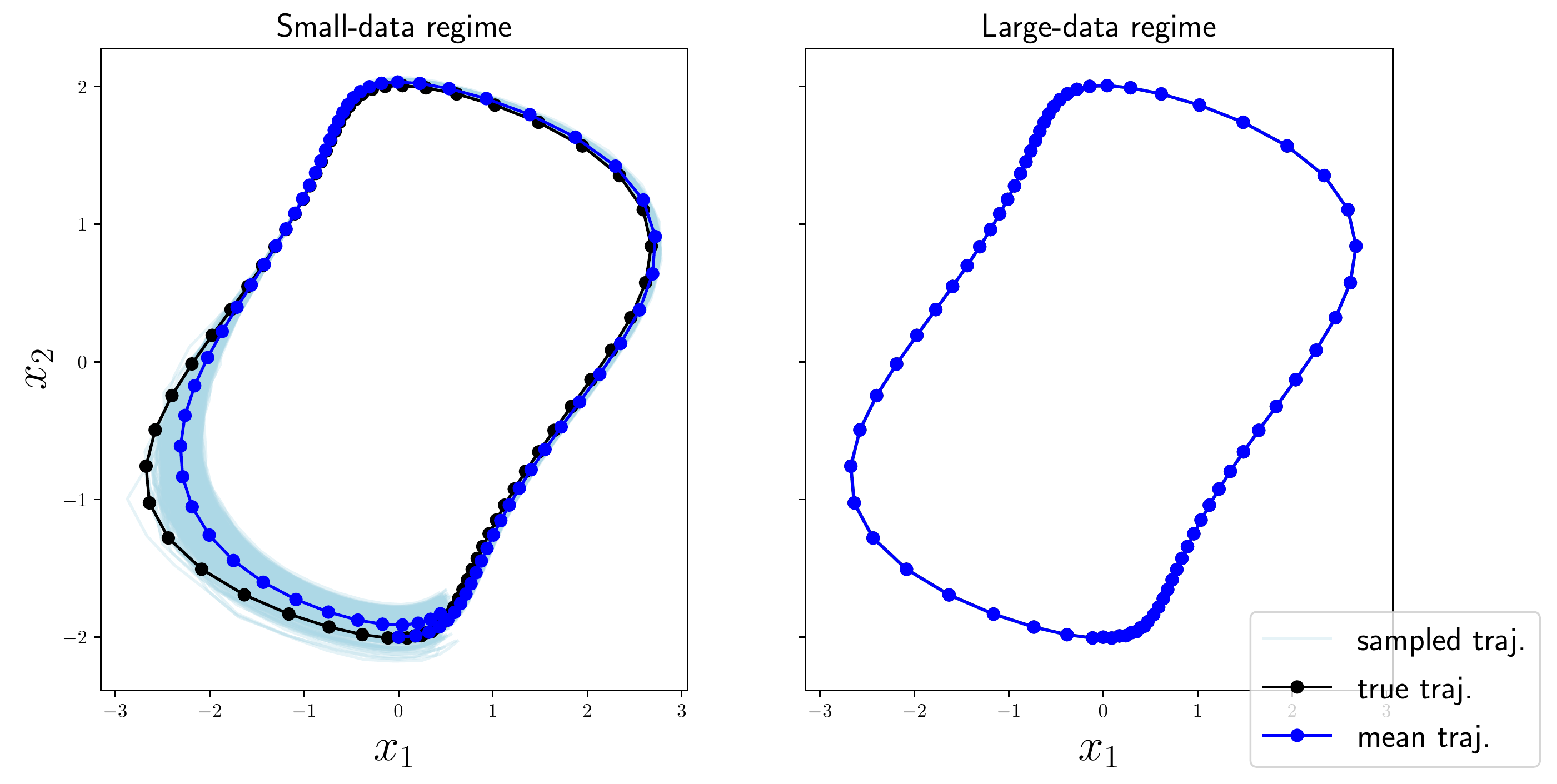}
    \caption{Learning the (nonlinear) Van der Pol oscillator with different amounts of training data. Our uncertainty-aware random feature model accurately learns the nonlinear dynamics with adequate data (\textbf{right}), while predicting with uncertainty under limited training data (\textbf{left}). Plausible sample trajectories (given a predefined confidence level) are depicted in light blue. Note that although the true trajectory sometimes deviates from the model (mean) prediction, it remains close to the sampled trajectories.}
    \label{fig:time_evo}
\end{figure}

In providing safety guarantees with GP dynamics, works such as \cite{kollerLearningbasedModelPredictive2018} assumed the true dynamics to live in an RKHS.
From a numerical analysis perspective, this is a reasonable assumption since the RKHS is dense in the space of continuous functions. 
Intuitively, RKHS functions can be used as powerful universal function approximators analogous to the celebrated \emph{Weierstrass approximation theorem}.
Motivated by that insight and the well-known connection between GP and kernel methods~\cite{kanagawaGaussianProcessesKernel2018}, this paper proposes to directly approximate the uncertain dynamics function using \emph{random features} (RF), a function approximation tool originated from large-scale kernel machines~\cite{rahimiRandomFeaturesLargescale2008}.
Due to its simple \emph{\lip} structure, RF have been extensively used in analyzing the theoretical foundation of statistical and deep learning, see, e.g., \cite{bachBreakingCurseDimensionality2017,belkinUnderstandDeepLearning2018,carratinoLearningSGDRandom2018,eeMachineLearningContinuous2020,hastieSurprisesHighDimensionalRidgeless2020,meiGeneralizationErrorRandom2020,rudiGeneralizationPropertiesLearning2017,sunApproximationPropertiesRandom2019}, and \cite{liuRandomFeaturesKernel2021} for a recent survey. Significant attention, such as the \emph{NeurIPS 2017 Test-of-Time Award}~\cite{rahimiRandomFeaturesLargescale2008} and \emph{ICML 2019 Honorable mentions}~\cite{liUnifiedAnalysisRandom2019}, have been dedicated to studies of random features.

In this paper, we exploit the function approximation capacity of RF for learning uncertainty-aware dynamics.
We summarize our \emph{contributions} and sketch the main results.
\begin{enumerate}[noitemsep,topsep=0pt]
	\item We propose the set-valued uncertainty-aware random feature (URF) dynamics model. Our model lifts the dynamics function to a reproducing kernel Hilbert space, allowing it to learn general nonlinear systems while quantifying the uncertainty due to limited observations (See Fig. \ref{fig:time_evo}).
	\item Exploiting the lifting structure of URF, we propose an indirect method to find the worst-case realization of the learned dynamics using \pmp, where the Hamiltonian is linear in the URF dynamics parameter.
	\item By viewing the one-step dynamics model as a shallow neural network, we take the perspective that the whole dynamical system can be viewed as a DNN (as well as a Bayesian net).
	We then show that, thanks to the URF model, our \pmp-based method is equivalent to the \cgd\ (also known as the \fw algorithm) on the DNN.
	\item Finally, we show that the eigenvalue-decay structure of common RKHSs allow us to learn a low dimensional representation of the dynamics model.
\end{enumerate}

The rest of the paper is organized as follows. 
In Section~\ref{sec:bg}, we briefly visit the background of kernel methods and random features.
In Section~\ref{sec:model}, we propose our main dynamics model --- the uncertainty-aware random feature model, and propose to find the worst-case dynamics via \pmp.
All proofs are deferred to Section~\ref{sec:pf}.
We then detail the concrete learning algorithms to estimate such models in Section~\ref{sec:learnSet}.
Our methodology is validated through numerical experiments on learning various dynamical systems in Section~\ref{sec:exp}.
Section~\ref{sec:related} contains other related works in the literature.
We conclude with a discussion in Section~\ref{sec:conclude}.

\section{Background}
\label{sec:bg}

In this paper, we assume a partially-known discrete-time dynamics according to\footnote{\textbf{Notation}: lower/upper case symbols denote vectors/matrices. We write scalars in both lower and upper case. Curly symbols (e.g., \wset) denote sets. $\nabla_x f(w)$ is the gradient of $f$ with respect to $x$ evaluated at $w$. $I_p$ denotes the $p \times p$ identity matrix. A normally distributed vector $x$ with mean $\mu$ and covariance $\Sigma$ is written as $x \sim \mathcal{N}(\mu, \Sigma)$. We use $\mathcal{U}(a,b)$ to denote the uniform distribution over the interval $(a,b)$. The euclidean norm of $x$ is denoted by $\| x \|_2$. We use $p(x|y)$ to refer to the conditional probability density of $x$ given $y$. $[M]_{i,j}$ denotes a specific entry of the matrix $M$.}
\begin{equation}
    \mathbf{x}_{t+1} = h(\mathbf{x}_t, \mathbf{u}_t) + f(\mathbf{x}_t, \mathbf{u}_t),
\label{eq:dynamics}
\end{equation}
where $\mathbf{x}_t \in \mathbb{R}^p, \mathbf{u}_t \in \mathbb{R}^q$ denote the system state and input action, respectively; $h(\cdot, \cdot)\colon \mathbb{R}^{p+q} \to \mathbb{R}^p$ represents a (known) nominal model and $f(\cdot, \cdot)\colon \mathbb{R}^{p+q} \to \mathbb{R}^p$ accounts for uncertain deviations from the nominal component. We are interested in learning the unknown term from collected transitions of a system governed by \eqref{eq:dynamics}.
Since this work focuses on dynamics learning instead of control design, we will suppress the control input $u_t$ henceforth.
Also, without loss of generality, we consider a single-output system $p=1$ throughout this work.
Note that we can easily handle multi-output systems by independently applying the proposed representation to each output dimension.

\subsection*{Learning with kernels and random features}
\newcommand{\rkhs}{\ensuremath{\mathcal H}\xspace}
Kernel-based machine learning uses smooth function spaces to learn the true data-generating functions in nature based on empirical data.
We refer interested readers to \cite{scholkopfLearningKernelsSupport2002,steinwartSupportVectorMachines2008}
for full coverage.
In the context of this paper, we focus on the regression setting, which can be viewed as the following approximation problem, also referred to as the risk minimization problem
$$
    \min_{f\in \mathcal H}\risk (f):= \int (f(x) - f^*(x))^2 \mu(d x),
$$
for some data-generating distribution $\mu$.
Given a data set $\mathcal{D} = \{(x_i,y_i)\}_{i=1}^T$, one can show using the representer theorem that the solution to the above variational problem admits a finite representation of the form
$
\hat f=\sum_{i=1}^T\alpha_i k(x_i,\cdot)
$,
where $k$ is a positive (semi-)definite kernel defined as a symmetric (real-valued) function, i.e., $\sum_{i=1}^T \sum_{j=1}^T a_i a_j k(x_i, x_j)\ge 0$ for any $T \in \mathbb{N}$, $\{ x_i \}_{i=1}^T \subset \mathcal{X} \subset \mathbb{R}^p$, and $\{a_i\}_{i=1}^T \subset \mathbb{R}$. 

One can also show that every positive definite kernel $k$ is associated with a Hilbert space \rkhs and a feature map $\phi\colon
\mathcal{X} \to \rkhs$, for which $k(x,y) = \langle \phi(x), \phi(y)
\rangle_\rkhs$ defines an inner product on $\rkhs$, where $\rkhs$ is a space of real-valued functions on $\mathcal{X}$. The space $\rkhs$ is called a reproducing kernel Hilbert space (RKHS), equipped with the \emph{reproducing property}: 
$f(x) = \langle f, \phi(x) \rangle_\rkhs$ for any $f\in \rkhs, x \in
\mathcal{X}$.
We denote the canonical feature map as $\phi(x):=k(x, \cdot)$.
One computational concern of kernel methods is that forming the Gram matrix $K$ ($[K]_{i,j} := k(x_i, x_j)$) and performing operations with it (e.g., matrix inversion) is expensive. To alleviate that burden, we now overview the RF approach.

The RF framework seeks to approximate the kernel function $k(\cdot, \cdot)$ (and the corresponding RKHS functions) through a randomized finite feature map $\hat{\phi}(\cdot)$ such that
\begin{equation}
	\label{eq:rf}
f(x) \approx {f}_\text{RF}(x)= w^\top \hat\phi(x), \ \ k(x, x')\approx
\sum_{i=1}^L \hat\phi_i(x)\hat\phi_i(x'),
\end{equation}
where $\{\hat\phi_i(x)\}_{i=1}^L$ are the random features.
Consider, for instance, random Fourier features (RFF)~\cite{rahimiRandomFeaturesLargescale2008} to approximate stationary (i.e., shift-invariant) kernels, where
$
\hat\phi_i(x) = \sqrt{2/L} \cos(a_i^{\top} x+b_i),
$
and the vector $a_i$ is sampled proportional to the kernel's spectral density and the offset as $b_i\sim \mathrm{Uniform}[0,2\pi]$. Interestingly, we can obtain an approximation to the popular Gaussian RBF kernel
$k (x,x')=e^{-{{\|x-x'\|}^2_2}/2l^2}$ by sampling
$a_i\sim \mathcal{N}(0, l^{-2}\mathbf{I}_p)$.

One strength of modern kernel methods is the richness of certain RKHSs.
Concretely, \emph{universal RKHSs}, e.g., that associated with the \rbfk, are dense in the space of continuous functions that vanish at infinity, cf. \cite{steinwartSupportVectorMachines2008,sriperumbudurUniversalityCharacteristicKernels2011}.
This density makes RKHSs an ideal choice for approximating functions.
Since RFs approximate features in the RKHS, they are also universal function approximators suitable for learning dynamics functions.

One prominent use of RF in recent machine learning literature is in studying the properties of DNN, such as in analyzing the so-called \emph{double descent} phenomenon~\cite{belkinUnderstandDeepLearning2018}.
This is due to the fact that RF can be seen as a two-layer shallow neural network. For example, instead of the Fourier features, one may choose to construct the random ReLU feature (see, e.g., \cite{belkinUnderstandDeepLearning2018,sunApproximationPropertiesRandom2019}) by choosing the feature as
$
\hat\phi_i(x) = \max (0, a_i  x + b_i).
$
Later in this paper, we consider the neural-network perspective by viewing the dynamical system across multiple time steps as a deep neural network.
We refer to the references in Section~\ref{sec:intro} and a recent survey \cite{liuRandomFeaturesKernel2021} for a detailed coverage of RF.

\section{Uncertainty-aware random feature dynamics model}
\label{sec:model}
We now propose our main set-valued  dynamics model, the \emph{uncertainty-aware random feature} (URF) dynamics model
\begin{multline}
    \mathbf{x}_{n+1} \in \mathcal F_\textrm{URF}(\mathbf{x}_n,\wset):=\\
    \bigg\{ h(\mathbf{x}_n) + \fxuw:\mathbf{w}_n\in\wset\bigg\}.
\label{eq:dynamicsRF}
\end{multline}
Using the terminology of robust optimization, $\wset$ is an uncertainty set, which can be data-driven and learned from data, or set by practitioners as a robustness tuning parameter.
URF considers the parameter to lie within some ellipsoid given by
\begin{equation}
	\wset := \{\mathbf{w} : (\mathbf{w} - \boldsymbol\mu)^\top \boldsymbol\Sigma^{-1} (\mathbf{w} - \boldsymbol\mu) \leq 1\},
	\label{eq:credible_interval_ddro}
\end{equation}
for some mean $\mu$ and covariance matrix $\Sigma$.
Alternatively, we also discuss how to learn a lower-dimensional representation of \wset in Section~\ref{sec:pca}.
In the case that the uncertainty set shrinks to a singleton estimate $\wset=\{\mu\}$, we refer to the resulting model as \emph{certainty-equivalent random feature} (CERF) dynamics model.

Previously, the authors of \cite{mohammadiEstimationUncertainARX2015} considered ellipsoidal parameter uncertainty sets with linear dynamics.
By contrast, we lift the dynamics into an approximate RKHS by modeling the uncertain (nonlinear) dynamics part using RF
\begin{equation}
	\label{eq:rfDef}
f(x,w) := \hat{\boldsymbol{\phi}}(\mathbf{x})^\top \mathbf{w},
\end{equation}
where $\hat\phi(\cdot)$ is the random feature map; e.g., random ReLU feature, random Fourier features.

\begin{remark}
	Our URF model is equivalent to a two-layer Bayesian neural network (BNN) with an ellipsoidal uncertainty set for the weights.
	However, compared with typical BNN designs such as those using dropout, our ellipsoidal uncertainty set has the advantage of tractable optimization as we shall show.
	
	Our model~\eqref{eq:dynamicsRF} can also be seen as a distributional robustness model~\cite{delageDistributionallyRobustOptimization2010}, by constraining the future state distribution in an ambiguity set 
	$$
	P_{x_{n+1}} \in \bigg\{ f(\cdot, w_n)_\sharp P_{x_n}\ :\ {w}_n\in\wset\bigg\},
	$$
	where $f(\cdot, w_n)_\sharp$ is a push-forward operator.
\end{remark}

\subsection{Worst-case dynamics via Pontryagin's minimum principle}
In this section, instead of deriving approximate uncertainty propagation for nonlinear dynamics, we focus on finding the worst-case (or analogously best-case) scenario with respect to a given cost function and its propagation through the system.
Since, after all, the worst-case scenario is the key to robust control design.
To that end, we now explain how to characterize the condition for the worst-case realizations of the URF dynamics.
Given some overall cost objective $J$ of interest, e.g., accumulated cost in an OCP, and a horizon $N$ (see \eqref{eq:ocpIntro}) we wish to find the worst-case realization of the dynamics by solving
\begin{multline}
	\label{eg:argmaxJ}
\ws =\\ \argmax_{w_n\in\wset, n=0\dots N-1} \Jws.
\end{multline}
In the next section, we propose the framework to find the worst-case URF dynamics by characterizing the optimality condition via Pontryagin's minimum principle (PMP).
Naturally, the worst-case cost ${ J(w^*_0, w^*_1, \dots, w^*_{N-1})}$ can be used to stress test a system or certify certain controller design.

Due to the nonlinear dynamics, solving for the \wc dynamics under URF model~\eqref{eq:dynamicsRF} is not a convex optimization problem. However, the shallow structure of URF, and the reproducing property in general, is to view the dynamics as linear in a lifted space, e.g., an RKHS.
We now characterize the optimality condition for the worst-case dynamics in \eqref{eg:argmaxJ} via the Pontryagin's minimum principle.

Let us denote, with a slight abuse of notation, the total \emph{negative} cost as $\hat{J}(w_0, w_1, \dots, w_{N-1})=\sum_{n=1}^N \hat{c}(x_n)$, where $\hat{c}(x_n) = - c(x_n)$ denotes the negative of certain stage cost $c(\cdot)$. We note that the maximization in \eqref{eg:argmaxJ} can equivalently be formulated as the minimization of $\hat{J}$. For what follows, we now define the control Hamiltonian function as
\[
H(x,p,w) := \hat{c}(x) +  p^\top f(x,w),
\]
with $p \in \mathbb{R}^p$. For simplicity of the derivation, we omit the known part $h(x)$ of the model since it is independent of the variable $w$.
\begin{prop}[PMP for worst-case dynamics]
	\label{thm:pmp}
    Suppose the \ws is the set of worst-case realizations of the uncertainty set \wset and \xs is the corresponding state trajectory.
    Then there exist the co-state variables $\ps$ that satisfy the adjoint equations
    \begin{align*}
        &p^*_{n} = \nabla_x H(x^*_n,p^*_{n+1},w^*_n), & &p^*_N = \nabla_x \hat{c}(x^*_N).
    \end{align*}
    Furthermore, the worst-case dynamics parameter \ws minimizes the Hamiltonian
    \begin{equation}
        \label{eq:maxHam}
        w_n^* = \argmin_{w\in\wset} H(x^*_n,p^*_{n+1},w), \text{ for } n=0,\dots,N-1.
    \end{equation}
\end{prop}
The PMP for \wc dynamics motivates us to find the \wc URF dynamics via an \emph{indirect method} of optimal control, as illustrated in \algpmp. 
\begin{algorithm}
	\SetAlgoLined
	Initialize $w_0, w_1, \dots, w_{N-1}$ (e.g., $w_i = \boldsymbol{\mu}$, see \eqref{eq:credible_interval_ddro})\\
	\For{$k = 0, 1, \dots$}{
		\textbf{Forward pass: shooting dynamics}\\
        Initialize $x_0$
        \\
		\For{$n = 0$ \KwTo $N-1$}{
			$x_{n+1} =  f(x_n,w_n)$ \\
		}
		\textbf{Backward pass: adjoint equation}\\
        Initialize $p_{N}=\nabla \hat{c}(x_{N})$\\
		\For{$n = N-1$ \KwTo $0$}{
			$p_{n} = \nabla_x H(x_n, p_{n+1}, w_n)$ \\
		}
        \textbf{Update \wc dynamics}\\
		\For{$n = 0$ \KwTo $N-1$}{
			Set
			$w_n = \argmin_{w\in\wset}
			{H}(x_n,p_{n+1},w)
			$\label{step:ham}\\
		}
	}
	\caption{PMP for \wc dynamics}\label{alg:pmp}
\end{algorithm} 

A few comments are in order.
    The Hamiltonian under URF dynamics is \lip, hence the objective of \eqref{eq:maxHam} is linear.
    Furthermore, Step~\ref{step:ham} of \algpmp can be performed incrementally.
    We will study the case where a conditional gradient descent step is used.
    Under URF, the state distribution is never explicitly propagated through the nonlinear dynamics, which would be generally intractable.
We now study the updating formula for the \wc dynamics in Step~\ref{step:ham} of \algpmp.
We write down the form of Hamiltonian minimization under URF dynamics
\begin{equation}
\label{eq:qclp}
\min_{w\in\wset}\Big\{H(x_n,p_{n+1},w) = \hat{c}(x_n) +  p_{n+1}^\top  \hat\phi(x_n)^\top w\Big\},
\end{equation}
where \wset is an ellipsoid defined by some $\mu$ and $\Sigma$, as in \eqref{eq:credible_interval_ddro}.
This optimization problem has a linear objective, a quadratic constraint in $w$, and can be efficiently solved by the following result.
\begin{prop}
	\label{thm:minHam}
	Under URF dynamics, the minimizer of the Hamiltonian in \eqref{eq:qclp} is given by
	$$
	w^*_n= \mu - \frac{\Sigma\hat\phi(x_n) p_{n+1} }{\sqrt{p_{n+1}^\top\hat\phi^\top(x_n) \Sigma\hat\phi(x_n) p_{n+1}}}, n=0,...,N-1.
	$$
\end{prop}

\begin{figure}[t!]
    \centering
    \includegraphics[width=0.6\columnwidth]{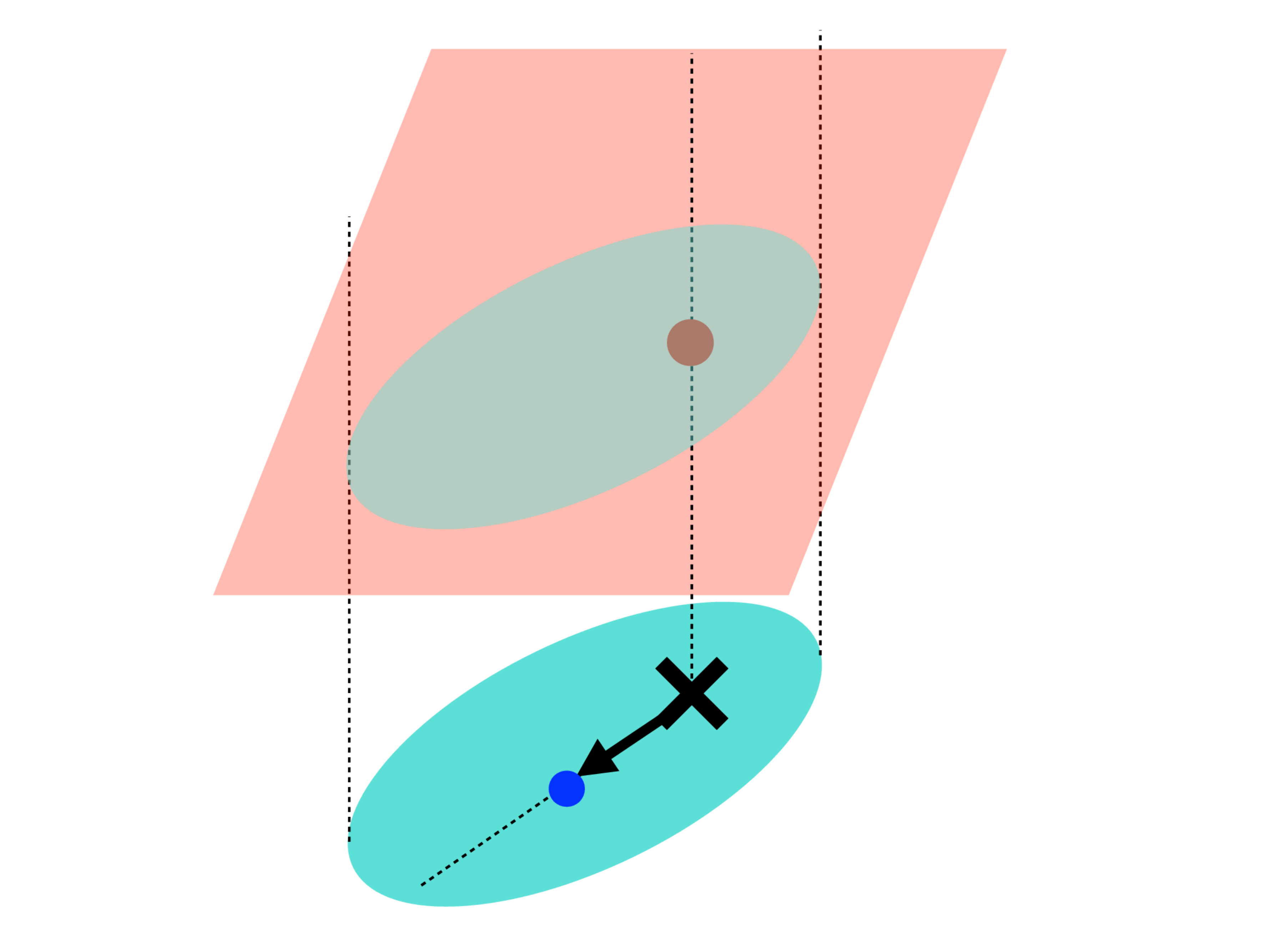}
    \put(-35,80){\small{${H}(x_n,p_{n+1},w_n)$}} 
    \put(-60,10){\small{\wset}} 
    \caption{Illustration of one step of the \fw algorithm, equivalent to inexact PMP. The Hamiltonian is depicted as a (salmon) plane considering its (affine) linear dependence on $w_n$. A particular uncertain set \wset (turquoise) is shown together with its induced constraint on the Hamiltonian values (grey). The black arrow goes from the current parameter value $w_n$ (cross) to the next value $w_n^+$ (blue dot) in the \fw minimization, as defined in \eqref{eq:gradHam} and \emph{equivalently} in \eqref{eq:fwDnn}. }
    \label{fig:fw}
\vspace{-0.5cm}
\end{figure}

\begin{remark}
    We can take the above argument beyond random features that are linear in parameters.
    By merit of the reproducing property of RKHSs, the Hamiltonian can be written as 
    \begin{equation}
    	\label{eq:hamRKHS}
        H(x,p,f) = \hat{c}(x) +  p^\top \hip{f}{\phi(x)},
    \end{equation}
    which is linear in the RKHS function $f$, see Fig.~\ref{fig:fw}.
    Hence, our approach can be also viewed as a (constrained) functional gradient approach where the decision variable is a RKHS function. Note that such functional gradients have been used in the recent kernel methods literature \cite{daiScalableKernelMethods2014,zhuKernelDistributionallyRobust2021} to optimize w.r.t. RKHS functions.
\end{remark}

\subsubsection*{Incremental update of Hamiltonian dynamics: equivalence to deep learning optimization}
An alternative way to perform the update in Step~\ref{step:ham} of \algpmp is to only incrementally optimize the Hamiltonian. For example, recent works such as \cite{liMaximumPrincipleBased2018} advocated for the incremental updates to avoid incurring large errors in the Hamiltonian dynamics for training DNN.
Enabled by the \lip structure of URF dynamics, we now propose to perform incremental Hamiltonian minimization
\begin{equation}
    \label{eq:gradHam}  
    \begin{aligned}
        \bar{w_{n}} &=& &\min_{w\in\wset} H(x_n,p_{n+1},w),\\
        {w_{n}^+} &=& &w_n + \gamma_k(\bar{w_{n}} - w_n),
    \end{aligned}
\end{equation}
for some step size schedule $\gamma_k$ and $n=0,...,N-1$.
We will refer to \algpmp with the incremental update~\eqref{eq:gradHam} as the \emph{inexact PMP}.

We now show that, the incremental minimization of Hamiltonian under URF dynamics is equivalent to performing the \cgd, also known as the \fw algorithm, on the weights of a DNN.
\begin{prop}[Equivalence of inexact PMP and Frank-Wolfe for deep learning]
    \label{thm:eqFrank}
    Inexact PMP,
    i.e.,
    \algpmp with update step replaced by \eqref{eq:gradHam},
    is equivalent to performing \fw algorithm on the total negative cost $\hat{J}$
    \begin{equation}
        \label{eq:fwDnn}
        \begin{aligned}
            \bar{w_{n}} &=& &\min_{w\in\wset} \gJ ^\top w,\\
            {w_{n}^+} &=& &w_n + \gamma_k(\bar{w_{n}} - w_n),
        \end{aligned}
    \end{equation}
Furthermore, \fw algorithm with $\gamma_k=1$ recovers the exact PMP solution.
\end{prop}

\begin{remark}
	Note that, if the dynamics is not \lip, the last statement of Proposition~\ref{thm:eqFrank} does not hold.
	That is the motivation of using PMP in the context of URF dynamics as a shallow representation.
	This can also be seen as a manifestation of the reproducing property in Hamiltonian dynamics~\eqref{eq:hamRKHS}.
\end{remark}
    To the best of our knowledge, \thmFw also constitutes a contribution to the current deep learning literature, aside from this paper's context of learning dynamics. 
    It characterizes the equivalence of constrained minimization of Hamiltonian via PMP and optimizing deep models via \cgd, which generalizes the results in \cite{liMaximumPrincipleBased2018,zhangYouOnlyPropagate2019}.
    Compared with those works, our focus is on finding the \wc dynamics and our result applies to constrained optimization.

In summary, Proposition~\ref{thm:eqFrank} tells us that, to compute the \wc dynamics, it suffices to perform \cgd\ on the DNN induced by the URF dynamics.
Note that we can likewise compute the best-case dynamics.
Furthermore, if we use code libraries based on computational graphs, such as PyTorch, the backward pass does not need to be implemented explicitly. We remark that the previous results assume a given uncertainty set $\wset$. Next we describe an approach to estimate $\wset$ in a data-driven fashion.

\section{Dynamics learning algorithms}
\label{sec:learnSet}
In this section we specify the details of learning the URF model~\eqref{eq:dynamicsRF} proposed in the previous section.
In addition, we provide a complementary dimensionality-reduction procedure based on nonlinear principal component analysis (PCA).

\subsection{Learning uncertainty set \wset via Bayesian linear regression (BLR)}
\label{ssec:blr}
We estimate the uncertainty set \wset in a data-driven fashion via BLR and ellipsoidal credible bounded regions. We assume access to observations of the full state of the system in $\eqref{eq:dynamics}$ together with a noisy version of the consecutive state, following common practice \cite{kollerLearningbasedModelPredictive2018,hewingCautiousModelPredictive2020}.
These observations form a dataset
$\mathcal{D} = \{(x_i, y_i)\}_{i=1}^T$
with
\begin{equation}
y_i = h(x_i) + f(x_i) + \epsilon_i,\quad \epsilon_i \sim \mathcal{N}(0,\sigma^2),
\label{eq:original_dataset}
\end{equation}
where $\epsilon_i$ accounts for iid Gaussian noise realizations. Using the proposed RF dynamics in \eqref{eq:rfDef}, we can rewrite \eqref{eq:original_dataset} as
\begin{equation}
\hat{y}_i = \hat{\boldsymbol{\phi}}(\mathbf{x}_i)^\top w + \epsilon_i,
\label{eq:rf_dataset}
\end{equation}
where in the l.h.s. we have now the residuals $\hat{y}_i = y_i - h(x_i)$ instead.
We highlight that a particular realization of the weights $w' \in \mathbb{R}^L$ induces in turn a \emph{deterministic} dynamics model $f(\cdot, w')$. 
Taking a Bayesian perspective, we assume a standard Gaussian prior over the feature weights $w \sim \mathcal{N}(0, \mathbf{I}_L)$, which can be shown to yield a $L$-dimensional GP approximation~\cite{lazaro-gredillaSparseSpectrumGaussian2010, rasmussenGaussianProcessesMachine2005}. 
Then \eqref{eq:rf_dataset} becomes a linear-Gaussian model in which posterior inference over $w$ is computationally tractable. In fact, conditioned on the observed data $\mathcal{D}$, the posterior over weights $p(w|\mathcal{D})$ is also Gaussian with parameters
\begin{equation}
\begin{split}
    \boldsymbol\mu_{\boldsymbol\omega | \mathcal{D}} &:= (\boldsymbol\Phi(X)^\top \boldsymbol\Phi(X) + \sigma^2 \mathbf{I}_L)^{-1} \boldsymbol\Phi(X)^\top \mathbf{y},\\
    \boldsymbol\Sigma_{\boldsymbol\omega | \mathcal{D}} &:= (\boldsymbol\Phi(X)^\top \boldsymbol\Phi(X) + \sigma^2 \mathbf{I}_L)^{-1} \sigma^2,
\end{split}
\label{eq:blr_posterior}
\end{equation}
where $\boldsymbol\Phi(X) \coloneqq [\hat{\boldsymbol\phi}(\mathbf{x}_1), \hat{\boldsymbol\phi}(\mathbf{x}_2),\dots, \hat{\boldsymbol\phi}(\mathbf{x}_T)]^\top$ denotes the RF evaluated at the training inputs.
Note that it is also possible to perform the posterior update incrementally as new data becomes available, i.e., with streaming data.

To construct the data-driven ellipsoidal set we rely on credible bounded regions,
within which the unknown $w$ value falls with high probability after seeing the data set $\mathcal{D}$. Formally, we define a posterior credible region for $w$ given a probability level $\alpha_w$ as
\begin{equation}
\wset =  \{\mathbf{w} \in \mathbb{R}^L : (\mathbf{w} - \boldsymbol\mu_{\boldsymbol\omega | \mathcal{D}})^\top \boldsymbol\Sigma_{\boldsymbol\omega | \mathcal{D}}^{-1} (\mathbf{w} - \boldsymbol\mu_{\boldsymbol\omega | \mathcal{D}}) \leq \chi_L^2(\alpha_{\boldsymbol\omega})\},
\label{eq:credible_interval}
\end{equation}
where $\chi_L^2(\cdot)$ is the quantile function of the chi-squared distribution with $L$ degrees of freedom.
Intuitively, \wset captures a high-density ellipsoidal region of the posterior $p(w|\mathcal{D})$ whose size is controlled through $\alpha_w$.
We have thus obtained a data-driven characterization of the uncertainty set \wset for the URF dynamics~\eqref{eq:dynamicsRF}.

\subsection{Lower-dimensional representation using random feature nonlinear component analysis}
\label{sec:pca}
\newcommand{\gramrf}{\ensuremath{\Phi(X) \Phi(X)^\top}\xspace}
It is well-known that, for some common kernels, the kernel Gram matrix $K$ has special eigenspectrum structure (see \cite{zhuGaussianRegressionOptimal1997,rasmussenGaussianProcessesMachine2005} for the classical results and \cite{belkinApproximationBeatsConcentration2018} for alternative characterizations).
For example, the eigenvalues of \rbfk Gram matrices decay at an exponential rate.
As the RFF approximate the feature maps associated with the RKHS of the \rbfk, we can expect the data Gram matrix \gramrf to have rapidly decaying eigenvalues.
Intuitively, this gives us the power to capture high-dimensional data in its lower dimensional (shallow) representation, such as by using kernel PCA.
We now show how to exploit this structure to learn a \emph{lower-dimensional representation} of the URF dynamics.

Given the RF representation of the data, we perform PCA on the Gram matrix \gramrf to obtain the lower-dimensional representation, which we denote as
$
\hat{\psi}(x): = P \hat{\phi}(x)
$,
where $P$ is a PCA projection matrix of size $\hat{L} \times L$ obtained, e.g., by performing singular value decomposition of the  Gram matrix. Note that $\hat{L}$ denotes the new dimension of the feature-based representation of the dynamics and therefore we choose it such that $\hat{L} \ll L$.
As a consequence, our URF dynamics in \eqref{eq:rfDef} can be alternatively constructed using the \emph{lower dimensional representation} as
$$
f_{\text{PCA}}(x,w) \coloneqq  \hat{\psi}(x)^\top w = \hat{\phi}(x)^\top P^\top w,
$$
where $w \in \mathbb{R}^{\hat{L}}$.
We highlight that the BLR algorithm described in section \ref{ssec:blr} can seamlessly incorporate the lower dimensional representation by using $\boldsymbol\Psi(X) \coloneqq [\hat{\boldsymbol\psi}(\mathbf{x}_1), \hat{\boldsymbol\psi}(\mathbf{x}_2),\dots, \hat{\boldsymbol\psi}(\mathbf{x}_T)]^\top$ instead of $\boldsymbol\Phi(X)$ in \eqref{eq:blr_posterior}.
A rigorous statistical analysis of the approximation error is beyond our current scope, for which we refer to \cite{lopez-pazRandomizedNonlinearComponent2014} and a comprehensive survey \cite[Section~5.3]{liuRandomFeaturesKernel2021}.
Note that it is also possible to perform PCA without explicitly forming the Gram matrix via tailored numerical methods.

\begin{figure}
    \subfloat[Source-spiral]{\includegraphics[width=0.33\linewidth]{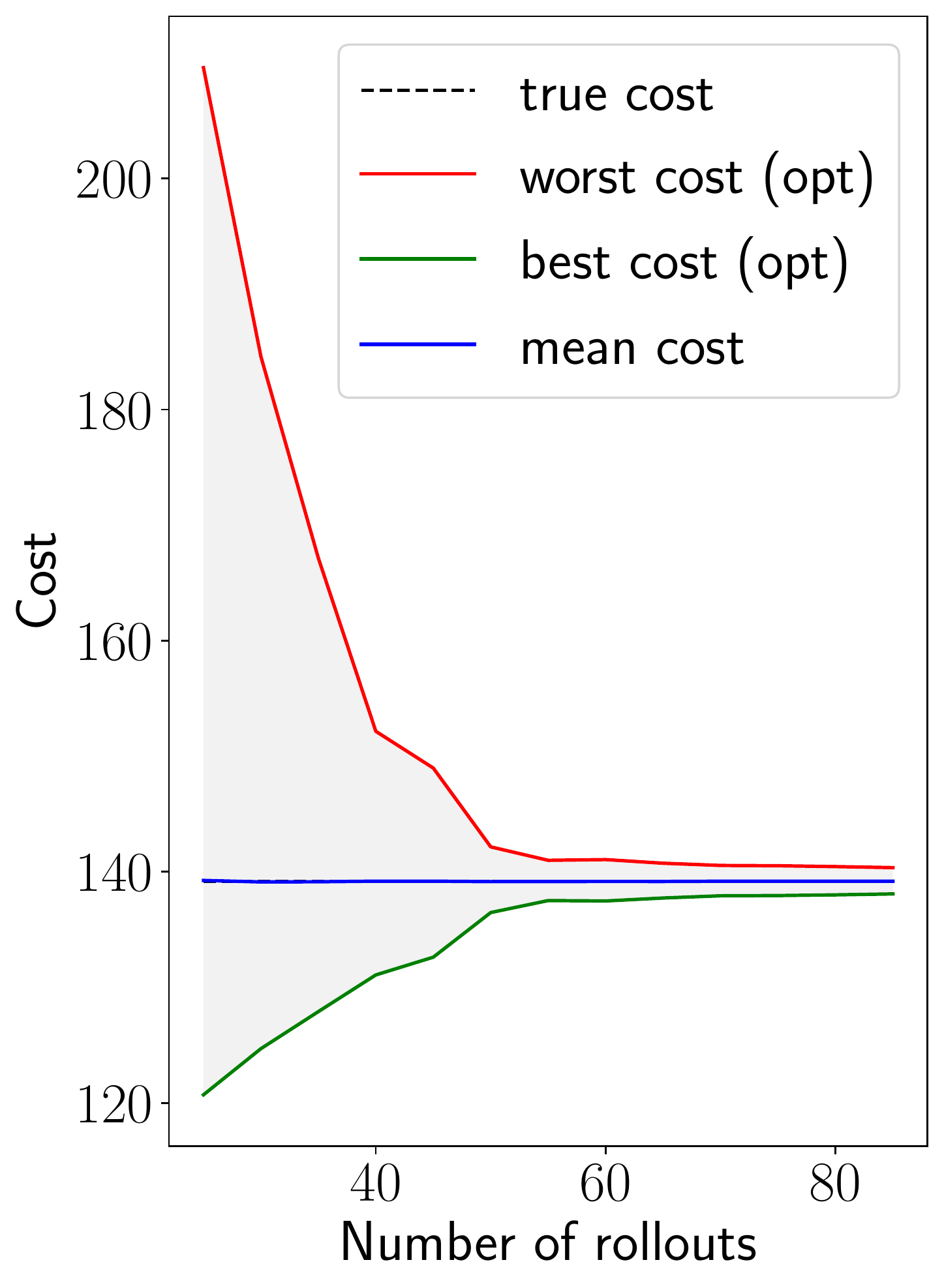}}
    \subfloat[Van der Pol]{\includegraphics[width=0.33\linewidth]{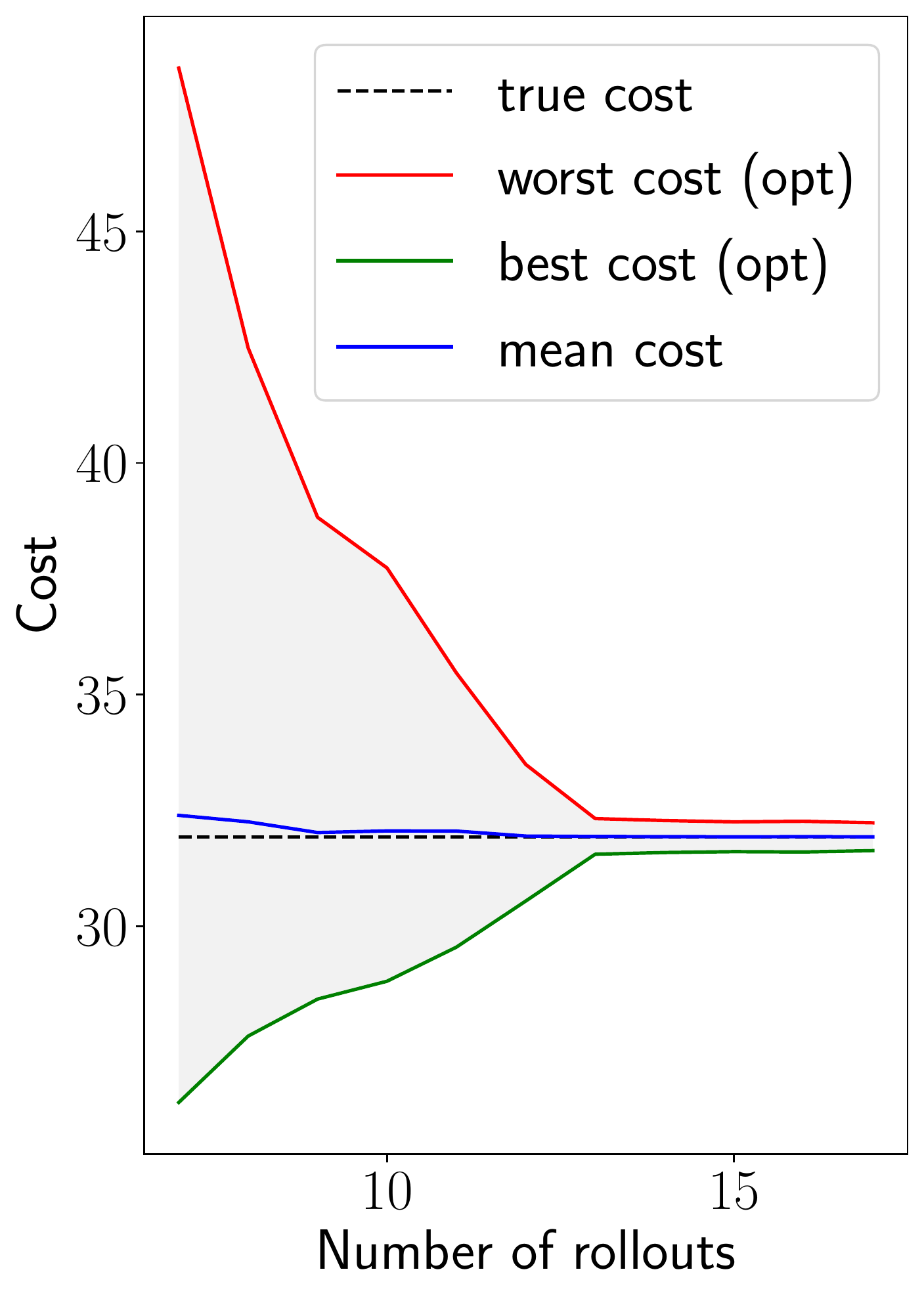}}
    \subfloat[Pendulum]{\includegraphics[width=0.33\linewidth]{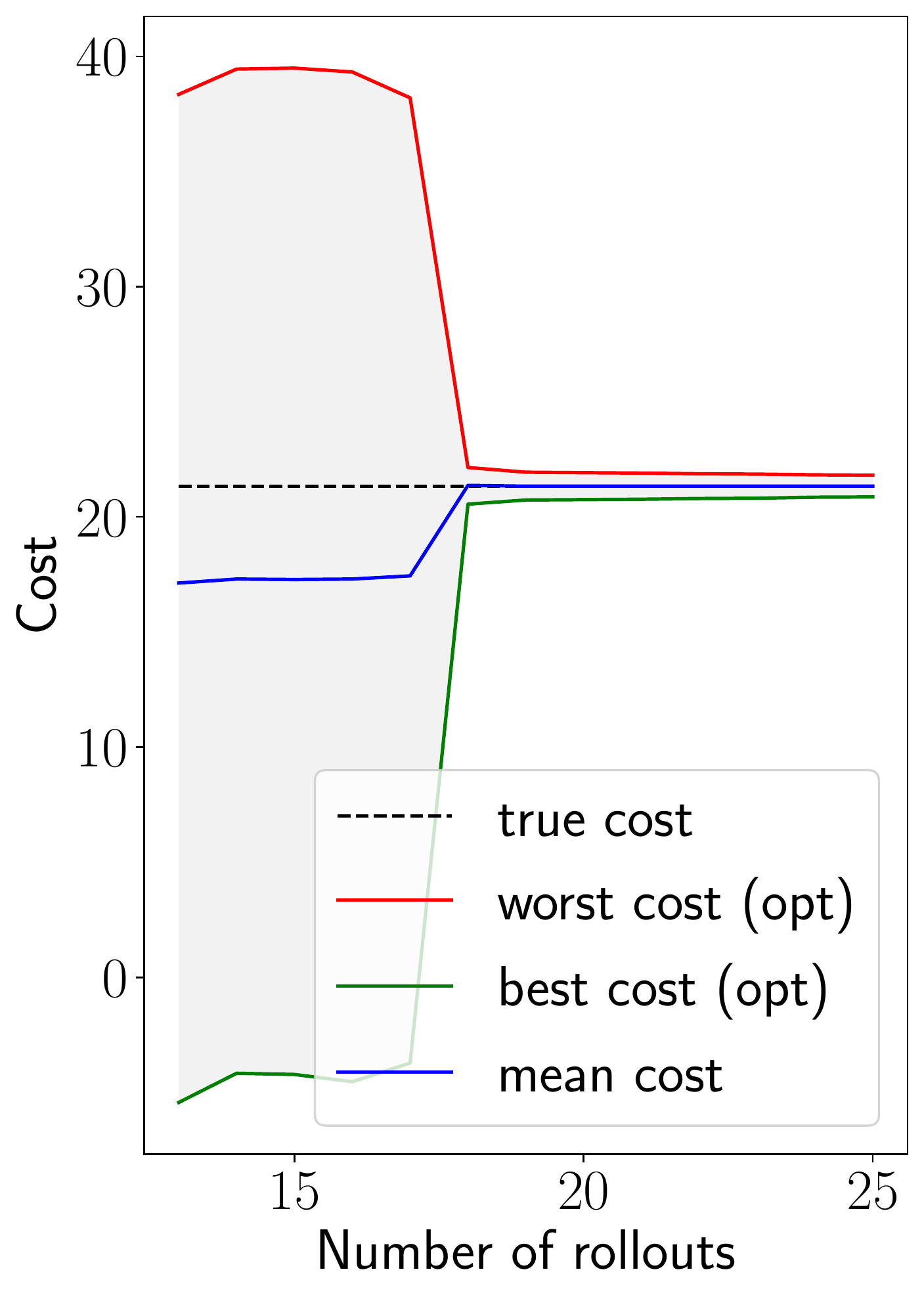}}
    \caption{Cost estimation of a fixed test trajectory (i.e., under a fixed initial condition) with the proposed URF model as a function of the number of training trajectories for three different systems. In addition to the mean model's prediction (blue) we depict the computed best (green) and worst (red) cost yielded by plausible dynamics within the inferred uncertainty sets (Algorithm \ref{alg:pmp}). Note that in the small-data regime the mean estimate might be biased, however the true cost is still contained in the range induced by the found best and worst values. The latter suggests that our URF model successfully captures the uncertainty due to limited training data. Both the uncertainty set and its induced best-worst range shrink as more training trajectories are used, converging to the true cost for large enough training datasets.}
    \label{fig:cost_fig}
\end{figure}

\section{Numerical experiment}
\label{sec:exp}

\begin{figure}
    \centering
    \includegraphics[width=\linewidth]{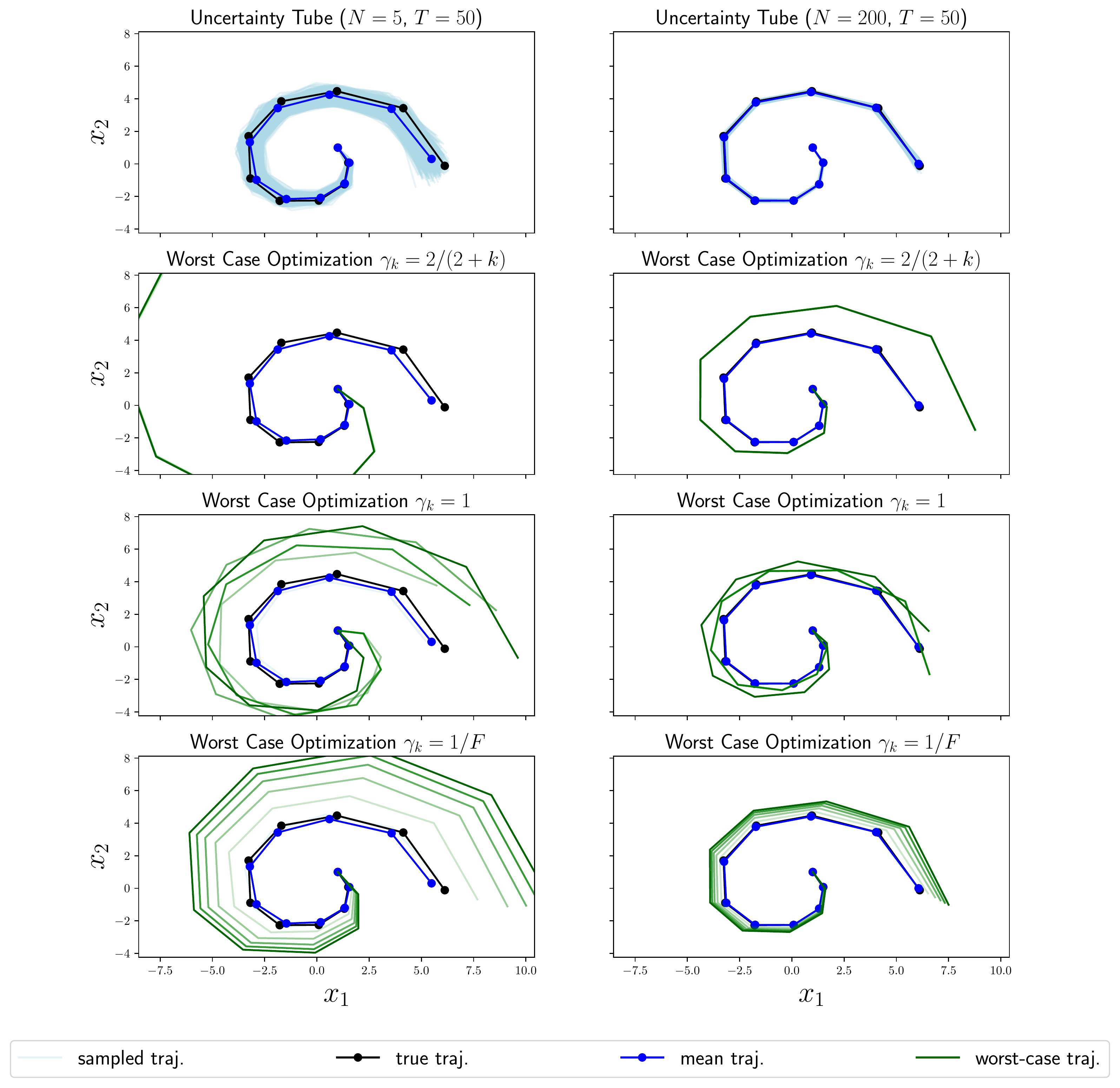}
    \caption{Worst-case dynamics optimization for different training data regimes using a nonlinear plant of the form $\mathbf{x}_{t+1} = \mathbf{A} \mathbf{x}_t + \cos (\mathbf{B}\mathbf{x}_t + \mathbf{c})$. The left column depicts the scarce training data setting ($N=5$ rollouts of length $T=50$ each) and the right column accounts for the large training data regime ($N=200$). \textbf{First row}: the predicted state mean trajectory (blue) is plotted alongside the true trajectory (black). Plausible state trajectories under the uncertainty set are also sampled uniformly and shown in light blue forming an \textit{uncertainty tube} around the mean prediction. \textbf{Bottom rows}: worst-case dynamics optimization for different schedulings of the learning rate $\gamma_k$ in \eqref{eq:gradHam}. Lighter green denotes earlier iterations of the optimization with the final result being shown in dark green. Remarkably, the \textit{uncertainty tube} shrinks as more training data is used for learning and the obtained worst-case trajectory is therefore closer to the mean trajectory for larger training datasets. Also note that the optimization is strongly dependent on the learning rate scheduling with the standard FW scheduling yielding better optimization performance in our experiments.
    }
    \label{fig:worst_dyn_opt}
\end{figure}

We illustrate the learning of URF dynamics in multiple autonomous discrete systems and showcase the possibility of finding the worst-case (and best-case) dynamics given a certain cost function. These examples enable us to study the influence of the size of the training sets in our data-driven characterization of the uncertainty. Furthermore, we explore different instances of the worst-case optimization procedure described in Algorithm \ref{alg:pmp}.

\begin{figure}
    \centering\subfloat[Van der Pol oscillator]{\includegraphics[width=0.8\linewidth]{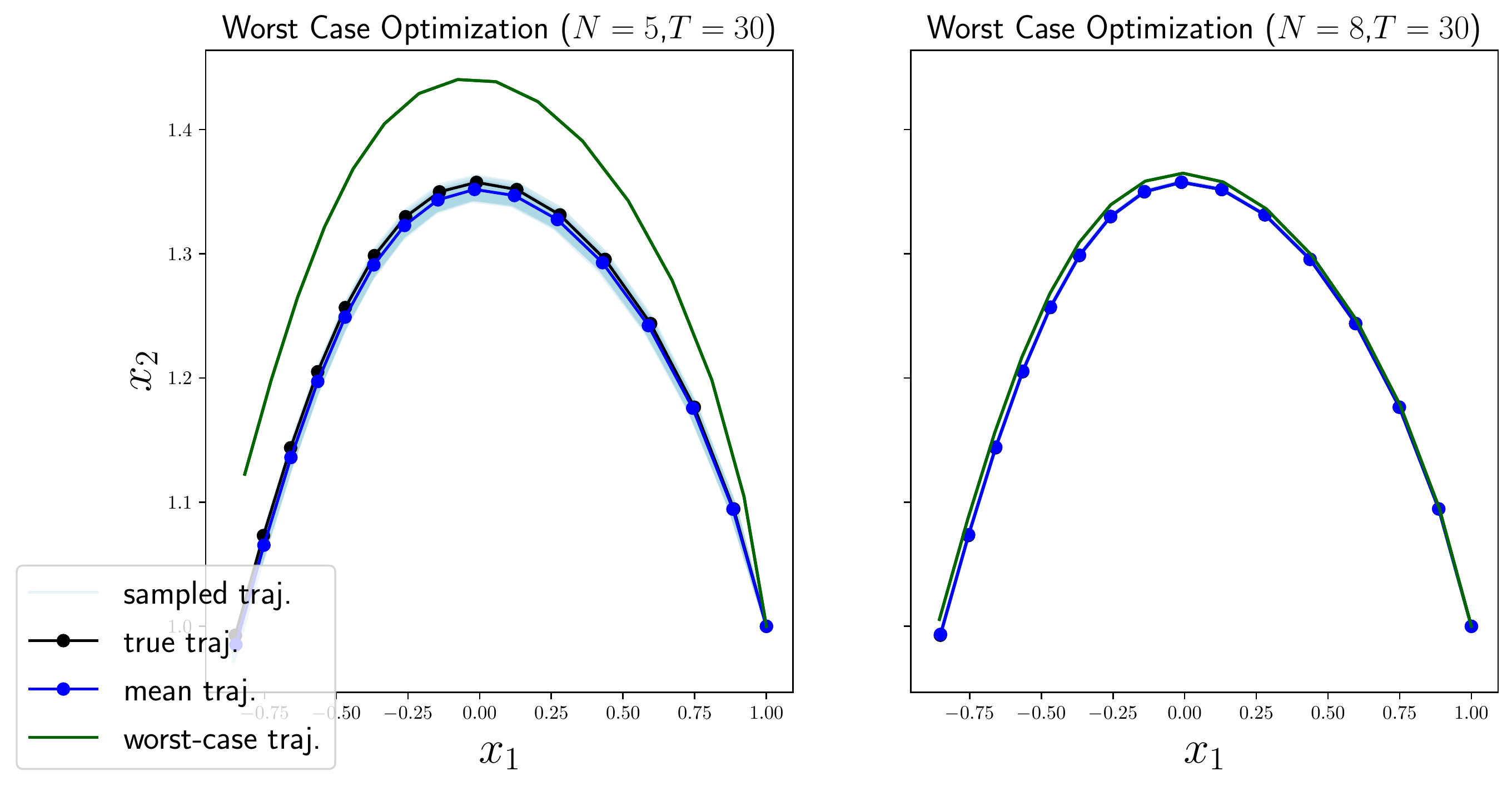}}\\
    \centering\subfloat[Damped Pendulum]{\includegraphics[width=0.73\linewidth]{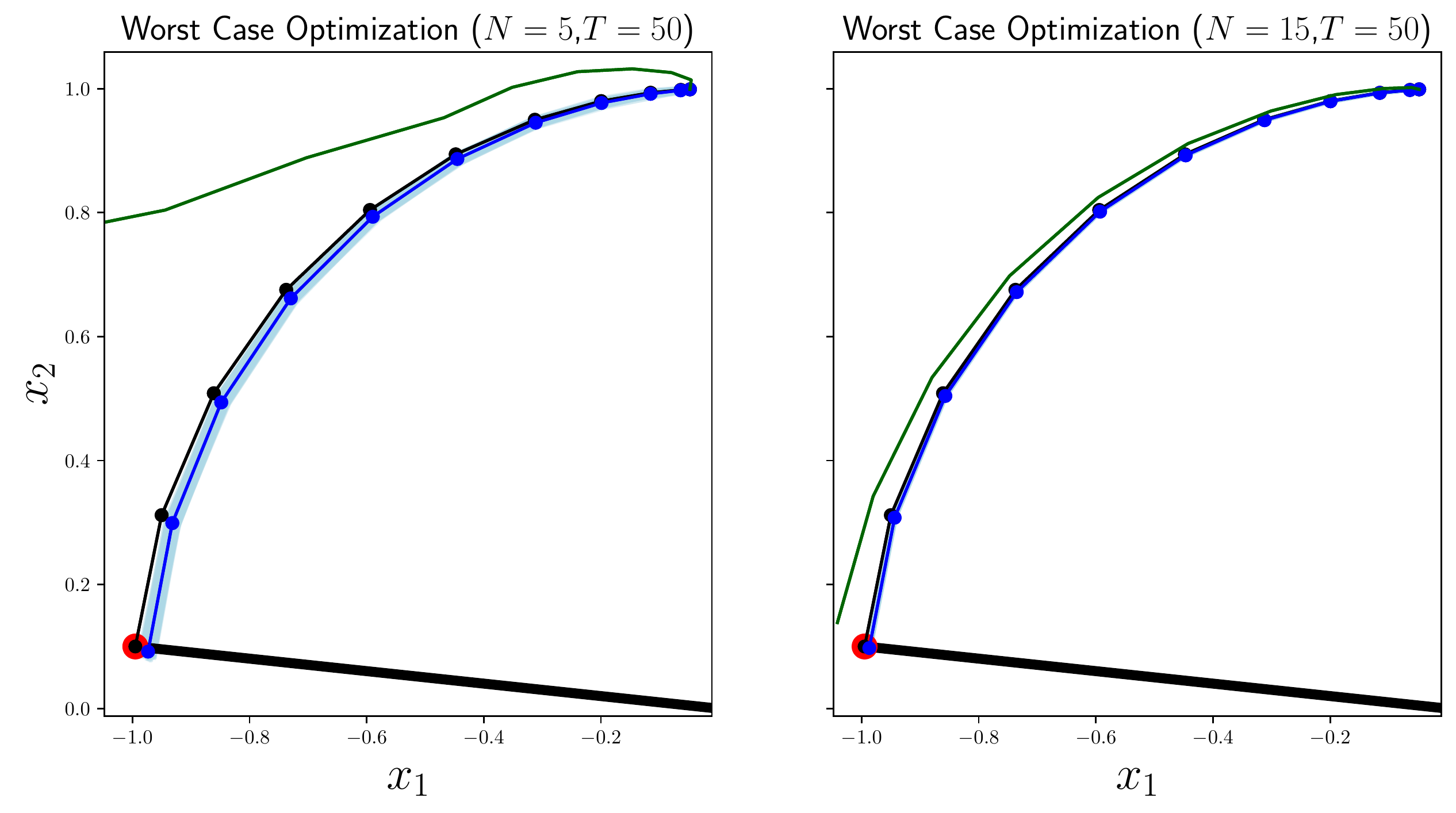}}
    \caption{Mean URF model prediction (blue) along with computed worst-case trajectory (green) for two nonlinear systems: the Van der Pol oscillator (\textbf{top row}) and the damped pendulum (\textbf{bottom row}). We consider both the scarce (\textbf{left column}) and large (\textbf{right column}) training data regimes and use the standard Frank-Wolfe learning rate scheduling for worst-case optimization. Note that under high uncertainty the worst-case trajectory might yield much higher cost than the true trajectory, whereas in the case of less uncertain model fits (i.e., large-data regime) even the worst case approaches the true trajectory.}
    \label{fig:pend_vdp}
\end{figure}

In all experiments we assume limited prior information about the underlying system and therefore our model consists of an identity nominal component $h(\mathbf{x}_t) = \mathbf{x}_t$. The residual $\mathbf{x}_{t+1} - h(\mathbf{x}_t)$ is then learned using the proposed URF model using as training data a varying number of randomly sampled trajectories ($N$) of fixed length ($T$). After model learning we can (i) predict future state trajectories given initial conditions, (ii) get an uncertainty \textit{tube} induced by the posterior distribution over the feature weights $\omega$ and (iii) find an adversarial realization of the state trajectories under such uncertainty for a particular task at hand; i.e., given a cost function $c(x_t)$ and a confidence level $\alpha_{\boldsymbol\omega}$. We depict graphically these capabilities in Figs. \ref{fig:worst_dyn_opt}, \ref{fig:pend_vdp} for three different nonlinear systems. In all our experiments we use $L=1000$ Fourier RF and apply RF nonlinear component analysis (Section \ref{sec:pca}) to obtain a lower-dimensional representation ($\hat{L}=100$).

First we consider a nonlinear plant of the form $\mathbf{x}_{t+1} = \mathbf{A} \mathbf{x}_t + \cos (\mathbf{B}\mathbf{x}_t + \mathbf{c})$, where $\mathbf{A}$ is fixed such that the resulting plot exhibits a source-spiral pattern as shown in Fig. \ref{fig:worst_dyn_opt}, and the affine mapping ($\mathbf{B}$,$\mathbf{c}$) is randomly sampled. We assume as cost function the quadratic form $c(x_t) = x_t^{\top} x_t$ and sample initial conditions according to $\mathbf{x}_0 \sim \mathcal{N}(0, \mathbf{I}_2)$.

We also learn a Van der Pol oscillator governed by the second-order differential equation $\dot{x}_1=(1 - x_2^2)x_1 - x_2, \dot{x}_2=x_1$ and use an explicit Runge-Kutta integrator for its discrete-time simulation. We sample initial conditions from $\mathcal{U}(-1,1)$ for training data generation. The URF model's prediction and obtained worst-case trajectory are shown in the top row of Fig. \ref{fig:pend_vdp} for the quadratic cost $c(x_t) = x_t^{\top} x_t$.

We finally consider a pendulum system under friction whose continuous-time dynamics is $\dot{x}_1=x_2, \dot{x}_2= -(g/l) \sin{x_1} -(\beta/(ml^2))x_2$, where the gravity is set to $g=9.81$; and the mass $m$, the rod length $l$ and the friction parameter $\beta$ are all set to 1. In training trajectories we sample the initial pendulum's angle from $\mathcal{U}(-\pi, \pi)$ and the initial angular velocity from $\mathcal{U}(-1, 1)$. We use semi-implicit Euler integration in order to get a discrete-time approximation of this system. To ease implementation we map the pendulum's state to a representation $\hat{x} = (a, b, c) = (l \cos x_1, l \sin x_1, x_2)$, where the pendulum's position is expressed in Cartesian coordinates, before learning the URF dynamics. The cost function is then defined as $c(\hat{x}) = b^2 -a + 0.1 c^2$, which encourages the pendulum to stay upright. Results are likewise graphically depicted in the bottom row of Fig. \ref{fig:pend_vdp}.

Moreover, we explore the role of the learning rate in the Hamiltonian maximization \eqref{eq:gradHam} for finding the worst plausible dynamics under a certain cost. In Fig. \ref{fig:worst_dyn_opt} we consider three variations in the scheduling of said learning rate in the Frank-Wolfe algorithm, namely the standard Frank-Wolfe scheduling $\gamma_k = 2 / (k + 2)$, $\gamma_k = 1$ (i.e., full steps) which implies that the optimization successively moves along the uncertainty set boundary and $\gamma_k = 1/F$, with $F$ being the total number of optimization steps. The optimization behavior is strongly dependent on the used learning rate schedule, with the standard Frank-Wolfe scheduling yielding the highest cost (i.e., the worse) dynamics for the performed experiments. 

We highlight that in the absence of large amounts of training data, the computed uncertainty set contains dynamics that could drive the system to high cost regions, as shown in the left columns of Figs. \ref{fig:worst_dyn_opt}, \ref{fig:pend_vdp}. However, once the inferred uncertainty set is reduced as a result of a larger training data set, the obtained worst-case dynamics optimization cost is closer to the cost incurred by the certainty-equivalent model (i.e., the mean model), which is consistent with the fact that in the limit of infinite data both worst and mean cost should converge to the same value since the uncertainty set will shrink to a singleton.

A finer-grained analysis of the previous point is presented in Fig. \ref{fig:cost_fig}, where we see that both the computed worst and best costs define a region (i.e., a tube) containing the true cost across different training data sizes and systems. Note that the gap between the worst, best, mean and true cost vanishes after enough training rollouts have been used. However, we emphasize that although we might get poor cost (mean) predictions under scarce training data (e.g., right column of Fig. \ref{fig:cost_fig}), the true cost is always within the region defined by the best- and worst-case curves. This hints that our method might be useful in robust control designs where the uncertainty from limited training data is taken into account.

\section{Other related works}
\label{sec:related}
Our use of \pmp for finding the \wc URF dynamics is similar to the use of adjoint sensitivity in robust nonlinear optimization \cite{diehlApproximationTechniqueRobust2006}.

Our shallow URF model is in contrast with a large body of deep learning literature that employ DNN as dynamics model, for which we refer to \cite{chuaDeepReinforcementLearning2018} for references.
As we have demonstrated, shallow URF models are rich enough to approximate general dynamics function while allowing the linear Hamiltonian structure.
Our perspective, which views the whole dynamical system as a DNN instead of the one-step dynamics model, is also consistent with recent mathematically principled analysis for deep learning via gradient flows, e.g., \cite{eeMachineLearningContinuous2020}.

The authors of \cite{zhangYouOnlyPropagate2019} used PMP for deep adversarial learning, but with no equivalence theorems for the constrained optimization.
Moreover, they used gradient projection algorithms which typically perform poorly in nonlinear programs. In \cite{kamtheDataefficientReinforcementLearning2018} a PMP-based analysis is presented for MPC with GP dynamics, however they rely on heuristic uncertainty propagation.

RF is closely related to the GP literature for dynamics learning.
The use of Fourier RF for general GP regression is known as Sparse Spectrum GP Regression \cite{lazaro-gredillaSparseSpectrumGaussian2010} and has been previously considered to learn dynamical system for filtering and control problems \cite{panPredictionUncertaintySparse2017}.
The authors of \cite{arcariMetaLearningMPC2020} used finite functional basis expansion for meta-learning, which is in the same spirit as RF.
RF has been also recently used to sample from GP dynamics \cite{hewingSimulationTrajectoryPrediction2020} and to devise more efficient general-purpose GP sampling algorithms~\cite{wilsonEfficientlySamplingFunctions2020}.

\section{Discussion and future work}
\label{sec:conclude}

In this paper we have proposed the uncertainty-aware random feature (URF) model for learning dynamical systems.
Our idea is to learn the one-step dynamics model as a shallow neural net, while the whole system can be seen as a deep net.
Exploiting the lifting structure of RKHS functions and RF, we propose a \pmp-based numerical algorithm to find the worst-case scenario of the dynamics model.
We further show that our approach is equivalent to performing the \cgd~(Franke-Wolfe) on a DNN.
Various numerical experiments validate the power of the proposed URF model.

In the future, we look forward to applying our dynamics model in data-driven robust control design, such as learning-based MPC. Another direction is to explore RF models in the direction of distributionally robust optimization and control. This is indeed possible as universal RKHSs have been shown to be effective tools for enforcing distributional robustness in optimization~\cite{zhuKernelDistributionallyRobust2021}.

\section{Proofs}
\label{sec:pf}
\subsection{Proof of Proposition~\ref{thm:pmp}}
\begin{proof}
The starting point is by viewing the dynamics as an \emph{adversarial player} in a two-player zero-sum game.
Then, the worst-case dynamics parameter $w_t$ is seen as the control input to the system dynamics $f(x_t,w_t)$.
Consequently , the proposition follows from the standard discrete-time Pontryagin's minimum principle; see, e.g., \cite[Volume~I, 4th Edition]{bertsekasDynamicProgrammingOptimal1995}.
\end{proof}
\subsection{Proof of Proposition~\ref{thm:minHam}}
In the following, we use $\Sigma^{1/2}$ to denote the Cholesky factor of the positive (semi-)definite matrix $\Sigma$.
\begin{proof}
We restate \eqref{eq:qclp} for convenience.
\begin{equation*}
    \begin{aligned}
        \min_{{(w-\mu)}^T\boldsymbol\Sigma^{-1}{(w-\mu)} \leq 1} 
        \hat{c}(x_n) +  p_{n+1}^\top  \hat{\phi}(x_n)^\top w.
    \end{aligned}
\end{equation*}
We use a change of variable $\Sigma^{1/2}v = w - \mu$, resulting in the optimization problem
\begin{equation*}
    \begin{aligned}
        \min_{{v}^Tv \leq 1} 
        \hat{c}(x_n) + \pphi\mu+ \pphisig v,
    \end{aligned}
\end{equation*}
where the first two terms are independent of the decision variable.
Then, we use the Cauchy-Schwarz inequality and the fact that ${v}^Tv \leq 1$ to write
$$
\pphisig  v\geq - \|\pphisigt\|_2,
$$
where the minimum is attained at
$$
v^* = - \frac{\pphisigt}{\|\pphisigt\|_2}.
$$
Hence, the minimizer of the Hamiltonian is given by
$$
w^* = \mu + \Sigma^{1/2}v^*  = \mu - \frac{\Sigma \hat{\phi}(x_n) p_{n+1} }{\|\pphisigt\|_2}.
$$
\end{proof}
\subsection{Proof of Proposition~\ref{thm:eqFrank}}
\begin{proof}
    It is an exercise (e.g., using \cite[Proposition~5]{liMaximumPrincipleBased2018}) to show
    \begin{equation}
        \label{eq:JeqH}
        \begin{aligned}
            \gJ &= \gHnw\\
            &= \gf p_{n+1}\\
            &= \hat{\phi}(x_n) p_{n+1}.
        \end{aligned}
    \end{equation}
The first step of \eqref{eq:fwDnn} has the closed-form solution
$$
\bar w_n = \mu - \frac{\Sigma\gJ }{\|(\Sigma^{1/2})^\top \gJ\|_2}.
$$
Plugging \eqref{eq:JeqH} into the above expression and noting Equation~\ref{eq:qclp}, we exactly recover \eqref{eq:gradHam}, which is the inexact PMP.

If $\gamma_k=1$, then the exact minimization of Hamiltonian is performed under linear dynamics. Hence, we recover exact PMP.
\end{proof}

\section*{Acknowledgment}
The authors thank Alexandra Gessner and Wittawat Jitkrittum for the helpful discussions. This work was supported by the German Federal Ministry of Education and Research (BMBF) through the Tübingen AI Center (FKZ: 01IS18039B).

\bibliographystyle{IEEEtran}
\bibliography{ref}

\end{document}